\documentclass[letterpaper,10pt,conference]{ieeeconf}
\IEEEoverridecommandlockouts
\usepackage{graphicx} 
\usepackage{booktabs}
\usepackage{amsfonts,amsmath,amssymb}
\usepackage{tikz}
\usetikzlibrary{arrows.meta,positioning}
\usetikzlibrary{calc}
\newcommand\inner[2]{\langle #1, #2 \rangle}
\title{Neural Operators for Design-Space Surrogate Modeling of \\Tendon-Actuated Continuum Robots}

\author{
    Branden Frieden$^{1}$,
    James M. Ferguson$^{1}$,
    Alan Kuntz$^{2}$, and 
    Varun Shankar$^{1}$
\thanks{$^{1}$The Robotics Center and the Kahlert School of Computing at the University of Utah, Salt Lake City, UT 84112, USA.}
\thanks{$^{2}$The Departments of Computer Science and Electrical and Computer Engineering at Vanderbilt University, Nashville, TN, 37203, USA.}
\thanks{}
\thanks{VS was supported by NSF DMS 2505986. Research reported in this publication was also supported by the Advanced Research Projects Agency for Health (ARPA-H) under Award Number D24AC00415-00 for ALISS. The ARPA-H award provided 25\% of total costs with an award total of up to \$10,741,534.20. The content is solely the responsibility of the authors and does not necessarily represent the official views of \mbox{ARPA-H}.}
}

\begin{document}

\maketitle

\begin{abstract}
Continuum robots enable dexterous manipulation in constrained environments, but require accurate and efficient models for real-time manipulation and control.
Traditional physics-based models can be computationally expensive and may suffer from inaccuracies due to unmodeled effects, while current learning-based methods often generalize poorly beyond the specific robot on which they are trained.
We present a formulation of surrogate modeling for tendon-driven continuum robots as an operator learning problem  that maps robot design parameters and tendon actuation inputs to resulting configurations. 
This formulation enables a single trained model to generalize across a large class of robot designs.
We develop four novel neural operator architectures--two based on Deep Operator Networks (DeepONets) and two based on Fourier Neural Operators (FNOs)--and train them on simulation data to predict robot configurations.
All architectures achieve good accuracy while allowing for fast and accurate generalization across designs. 
Our results demonstrate that operator learning provides an effective and generalizable surrogate for continuum robot mechanics in the design space, enabling fast modeling for control, planning, and design optimization in surgical and industrial applications.
\end{abstract} 

\section{Introduction}
\label{sec:intro}

Continuum robots (CRs) are manipulators with highly deformable backbones, capable of smooth, continuous motion rather than discrete jointed articulation. Their inherent compliance makes them well-suited for applications in minimally invasive surgery, inspection, and confined-space manipulation~\cite{Walker2013ContinuousBackbone,Webster2010ConstantCurvature,Dupont2022_ProcIEEE,Burgner2015_TRO}. Surveys highlight the breadth of CR designs and control approaches, emphasizing tendon-driven continuum robots (TDCRs) as a widely used actuation paradigm~\cite{Zhang2022survey,Russo2023continuumOverview,Dupont2022_ProcIEEE,Burgner2015_TRO}. Tendon actuation enables compact, dexterous designs, but introduces nonlinearities from friction, coupling, and backbone elasticity~\cite{Rao2021HowToModelTDCR,Rucker2011StaticsTendons}.

Physics-based mechanics models of CRs typically use Cosserat rod theory, which treats the backbone as an geometric three dimensional structure rod~\cite{Antman2005Problems,Rucker2011StaticsTendons}. 
This framework captures bending, torsion, shear, and extension, and forms the foundation of many practical applications. 
However, numerical integration of these models, especially when coupled with tendon routing and external loads, can be computationally demanding \cite{Rao2021HowToModelTDCR, Till2019RealtimeCosserat} and may not be as expressive as learned approaches which can capture unmodeled effects (e.g. friction).

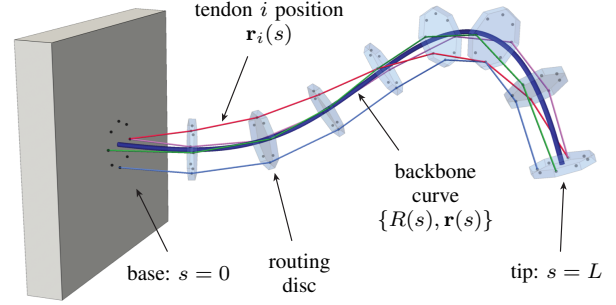
\begin{figure}
    \centering
    \begin{tikzpicture}[scale=1.0, every node/.style={font=\footnotesize}]
        \node[inner sep=0pt] (img) {\includegraphics[width=0.9\linewidth]{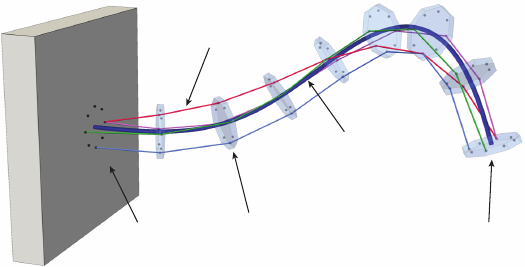}};

        \node[align=center, inner sep=2pt] at (1.7,-0.6) {backbone \\ curve \\ $\{R(s), \mathbf{r}(s)\}$};
        \node[align=center, inner sep=2pt] at (-0.5,1.7) {tendon $i$ position \\ $\mathbf{r}_i(s)$};
        \node[align=center, inner sep=2pt] at (-0.1,-1.6) {routing \\ disc};
        \node[align=center, inner sep=2pt] at (3.3,-1.6) {tip: $s=L$};
        \node[align=center, inner sep=2pt] at (-1.7,-1.6) {base: $s=0$};
    \end{tikzpicture}
    \caption{Tendon-driven continuum robot (TDCR) that we model with neural Operators. Tendons are routed through holes in the attached discs at a constant pitch $\theta$, and the flexible backbone bends continuously. Given tendon tensions, and robot design parameters, our model predicts the shape of the robot at any position along the backbone.}
    \label{fig:tdcr}
\end{figure}

A promising recent paradigm for overcoming these challenges is in learned or surrogate models of CRs~\cite{Kuntz2020_TMRB}. 
Prior work in soft and continuum robotics has emphasized the importance of efficient models for real-time control~\cite{Duriez2013RTFEM,Thuruthel2018ControlStrategies,DellaSantina2021Survey}. 
These modern methods allow models to \textit{learn} kinematic or static mappings directly from data, either generated in simulation via physics-based models or via sensing of the physical robot as it is actuated. 
Examples include surveys of machine learning–based control~\cite{Wang2021MLControlSurvey}, data-driven methods to compute the inverse kinematics~\cite{Xu2017_IJMRCAS, Liang2021_ICRA} and forward kinematics~\cite{Bergeles2015_Hamlyn, Grassmann2018_IROS,Kasaei2023_ICRA,Fagogenis2016_IROS,Kasaei2023_CoRL} of continuum robots, deep visual inverse kinematics for static shape control~\cite{Almanzor2023TRO}, model-less visual control of TDCRs using recurrent neurodynamic optimization~\cite{He2024RAS}, tip-force estimation using recurrent neural networks~\cite{Feng2021SciRep}, and deep decoder approaches to hysteresis compensation~\cite{cho_accounting_2024}. 
While these approaches are typically efficient and accurate, they are generally \emph{design specific}.
That is, if a robot’s tendon routing, disk spacing, or material properties change, new training data must be generated, either in simulation or by fabricating and actuating the physical robot.

In parallel, the scientific machine learning community has developed \emph{neural operators}, architectures that learn mappings between infinite-dimensional function spaces rather than finite-dimensional vectors~\cite{Kovachki2023NeuralOperator}. 
Deep Operator Networks (DeepONets)~\cite{Lu2021DeepONet} and Fourier Neural Operators (FNOs)~\cite{Li2021FNO} are leading frameworks, with demonstrated success in modeling complex physical systems including fluid dynamics, materials modeling, and climate forecasting~\cite{Li2021FNO,You2022IFNO,Pathak2022FourCastNet}. 
Transformer-based neural operators have also begun to appear, for example ViTO, which applies vision transformers to PDE surrogates~\cite{Ovadia2023ViTO}. 
Neural operators combine superior generalization capabilities with fast inference, making them natural candidates for surrogate modeling in a wide variety of applications. Operator learning is also beginning to impact robotics. Recent examples include a neural-operator predictor for delay-compensating control with closed-loop stability guarantees and experiments on a 5-link manipulator~\cite{Bhan2025PredictorNO}; an operator formulation of motion planning that learns a cost-to-value map via the Planning Neural Operator (PNO), enabling zero-shot super-resolution and generalization across environments~\cite{Matada2025PNO}; and neural-operator-based haptic sensing along slender rods, relevant to interventional and continuum-robot settings~\cite{Tang2025HapticNO}.

Inspired by these developments, we develop \emph{design-agnostic} surrogate models for TDCRs using two foundational neural operators: DeepONets and FNOs. To facilitate the use of neural operators as surrogates, we develop a mathematical formulation of TDCR surrogate learning as an operator learning task. Using data generated from Cosserat rod simulations of multiple designs with tendon actuation, we design and train custom DeepONets and FNOs to approximate the mapping from the design space of a tendon-actuated continuum robot to its equilibrium configuration space. This results in neural operator architectures that are able to output both positions in $\mathbb{R}^3$ and rotations in $SO(3)$. Our results show that these architectures achieve high accuracy on inference for unseen designs and fast evaluation times, demonstrating the potential of operator learning for design-agnostic modeling of continuum robots. 

The remainder of this paper is organized as follows. In Section \ref{sec:background}, we present a review of the Cosserat rod model and of operator learning. Next, in Section \ref{sec:methods}, we present a mathematical formulation of surrogate modeling of TDCRs as an operator learning problem, and present four new neural operator architectures (two based on DeepONets and two based on FNOs) suited to this task. We also discuss loss functions, important architectural details (for reproduction), and training details (including data generation). Then, in Section \ref{sec:results}, we present convergence results for our model on the task of learning equilibrium configurations, explore the parsimony (or lack thereof) of our models, the behavior of these new models on out-of-distribution predictions, and also present both training and evaluation timings. We conclude in Section \ref{sec:conclusion} with a summary and a description of future work.
\section{Background}
\label{sec:background}

\subsection{Continuum Robots as Cosserat Rods}
\label{sec:cosserat}

We briefly review the physics-based mechanics model commonly used for tendon-actuated continuum robots, in which the backbone is modeled as a geometrically exact Cosserat rod \cite{Antman2005Problems, rucker_statics_2011}. 
The shape of the rod is parameterized by the arc-length coordinate $s\in[0,L]$. 
Specifically, the centerline of the backbone is represented by a space curve $\mathbf r(s)\in\mathbb R^3$ together with an attached orthonormal frame $R(s)\in SO(3)$ describing the orientation of the cross-section; see Fig.~\ref{fig:tdcr} for an illustration.

The kinematic variables evolve along the rod according to
\begin{align}
\boldsymbol\nu(s)=R(s)^\top \mathbf r'(s),\qquad
R'(s)=R(s)\widehat{\boldsymbol\kappa}(s),
\end{align}
where $\boldsymbol\nu(s)$ and $\boldsymbol\kappa(s)$ denote the shear/extension and bending/torsion strain vectors, respectively.

Static equilibrium of the rod is governed by the balance of internal force $\mathbf n(s)$ and moment $\mathbf m(s)$ \cite{rucker_statics_2011}:
\begin{align}
\mathbf n'(s)+\mathbf f(s)&=0, \label{eq:force-balance}\\
\mathbf m'(s)+\mathbf r'(s)\times \mathbf n(s)+\boldsymbol\ell(s)&=0, \label{eq:moment-balance}
\end{align}
where $\mathbf f$ and $\boldsymbol\ell$ denote distributed forces and moments acting along the rod.
The stresses and strains inside the rod are related by the material constitutive law:
\begin{align}
\mathbf n = K_s(\boldsymbol\nu-\boldsymbol\nu_0),\qquad 
\mathbf m = K_b(\boldsymbol\kappa-\boldsymbol\kappa_0),
\end{align}
where $K_s, K_b$ are stiffness matrices encoding sensitivity to shear/elongation and bending/torsion, which can be computed from material properties.

Tendons routed at fixed cross-sectional offsets $\boldsymbol\rho_i$ generate distributed loads along the rod, resulting the shape deformation. 
The path of the $i$th tendon is given by
\begin{align}\label{eq:tend-path}
\mathbf r_i(s) = \mathbf r(s)+R(s)\boldsymbol\rho_i,
\end{align}
with unit tangent $\mathbf t_i=\tfrac{\mathbf r_i'}{\|\mathbf r_i'\|}$. 
The tendon tension $\tau_i$, causes distributed loads along the rod according to \cite{rucker_statics_2011}:
\begin{align}
\mathbf f_i(s)=-\frac{d}{ds}\big(\tau_i \mathbf t_i\big),\qquad
\boldsymbol\ell_i(s)=R(s)\,\boldsymbol\rho_i\times \mathbf f_i(s),
\end{align}
which contribute to the toal loads in\eqref{eq:force-balance}–\eqref{eq:moment-balance}~\cite{Rao2021HowToModelTDCR, rucker_statics_2011}.

Together, these equations describes the static equilibrium of the coupled rod–tendon system under prescribed base conditions and tendon actuation.
Solving the resulting boundary value problem yeilds the robot configuration.
The resulting system of equations serves as the high-fidelity physics model underpinning many continuum-robot simulators and controllers~\cite{Webster2010ConstantCurvature,Rao2021HowToModelTDCR,Zhang2022survey,Russo2023continuumOverview}. Consequently, in this work, we use this formulation as our ground truth representation of a tendon-actuated continuum robot.


\subsection{Operator Learning}
\label{sec:op-learn}
We now provide a brief mathematical background on operator learning. Let $\mathcal{U}\left(\Omega_u; \mathbb{R}^{d_u} \right)$ and $\mathcal{V}\left(\Omega_v; \mathbb{R}^{d_v} \right)$ be two separable Banach spaces of  $\mathbb{R}^{d_u}$- and $\mathbb{R}^{d_v}$-valued functions, respectively, taking values in $\Omega_u \subset \mathbb{R}^{d_x}$ and $\Omega_v \subset \mathbb{R}^{d_y}$, respectively. Further, let $\mathcal{G}: \mathcal{U} \rightarrow \mathcal{V}$ be a general (nonlinear) operator. The operator learning problem involves approximating $\mathcal{G} : \mathcal{U} \to \mathcal{V}$ with a parametrized operator $\hat{\mathcal{G}} : \mathcal{U} \times \Theta \to \mathcal{V}$  from a finite number of function pairs $\{(u_i, v_i)\}$, $i=1,\ldots,N$ where $u_i \in \mathcal{U}$ are typically called \emph{input functions}, and $v_i \in \mathcal{V}$ are called \emph{output functions}, \emph{i.e.}, $v_i = \mathcal{G}(u_i)$. The parameters $\Theta$ are chosen to minimize $\|\mathcal{G} - \hat{\mathcal{G}}\|$ in some norm. In practice, the problem must be discretized. This requires samples of the input and output functions at a finite set of function sample locations $X \in \Omega_u$ and $Y \in \Omega_v$, respectively. One then requires that $\|v_i(y) - \hat{\mathcal{G}}(u_i)(y)\|^2_2$ is minimized over $(u_i,v_i)$, $i=1,\ldots,N$, where $u_i$ are sampled at $x \in X$ and $v_i$ at $y \in Y$, for some chosen architecture for $\hat{\mathcal{G}}$. In the context of surrogate modeling for partial differential equations (PDEs), the spaces $\mathcal{U}$ and $\mathcal{V}$ often correspond to spaces initial and final conditions (in the case of time-dependent PDEs), all of which are functions, and the sets $X$ and $Y$ correspond to spatial sampling locations. DeepONets and FNOs correspond to specific architectural choices for $\hat{\mathcal{G}}$.

\section{Neural Operators as Surrogate Models for TDCRs}
\label{sec:methods}
\subsection{Operator Learning Formulation}
\label{sec:ol-cosserat}
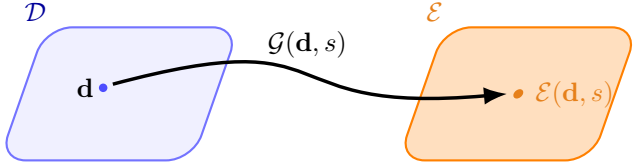
\begin{figure}[!htpb]
\centering
\begin{tikzpicture}[>=Latex,thick,scale=0.8]
  \def\W{3.2}\def\H{2.2}\def\R{10pt}\def\SL{0.35}\def\DX{6.6}

  \begin{scope}[xslant=\SL]
    \path[fill=blue!6,draw=blue!55,rounded corners=\R]
      (-1.8,-1.2) rectangle +(\W,\H);
  \end{scope}
  \node[blue!60!black] at (-1.8+0.18, -1.2+\H+0.25) {$\mathcal D$};
  \coordinate (d) at (-0.50,0.00);
  \fill[blue!70] (d) circle (2.0pt);
  \node[anchor=west] at ($(d)-(0.60,0)$) {$\mathbf d$};

  \begin{scope}[shift={(\DX,0)},xslant=\SL]
    \coordinate (Emin) at (-1.8,-1.2);
    \coordinate (Emax) at ($(Emin)+(\W,\H)$);
    \path[fill=orange!25,draw=orange,rounded corners=\R]
      (Emin) rectangle (Emax);
   \path let \p1 = (Emin), \p2 = (Emax) in
  coordinate (Ecenter) at ($(\p1)!.5!(\p2)$);
    \fill[orange!60!brown] (Ecenter) circle (2.2pt);
  \end{scope}
  \node[orange!60!brown] at (\DX-1.8+0.18, -1.2+\H+0.25) {$\mathcal E$};
  \node[anchor=west,orange!60!brown] at ($(Ecenter)+(0.12,0)$) {$\mathcal E(\mathbf d,s)$};

  \draw[->,line width=1.3pt]
    ($(d)+(0.16,0.06)$)
      to[out=15,in=185,looseness=1.6]
    node[midway,above,yshift=2pt] {$\mathcal G(\mathbf d,s)$}
    ($(Ecenter)+(-0.18,0)$);
\end{tikzpicture}
\vspace{-0.6em}
\caption{Surrogate modeling of tendon-driven continuum robots as an operator learning problem. A design $\mathbf d\in\mathcal D$ (blue) is mapped by the operator $\mathcal G(\mathbf d,s)$ 
to its equilibrium $\mathcal E(\mathbf d,s)\in\mathcal E$ (orange).}
\label{fig:ol-pic}
\end{figure}
We now formulate the Cosserat rod model as an \emph{operator} that maps from a \emph{design space} $\mathcal{D}$ to a corresponding \emph{space of equilibrium configurations} $\mathcal{E}$ for a given fixed initial configuration of the rod. This differs from the standard setup of operator learning in Section \ref{sec:op-learn} in one crucial way: while the elements of $\mathcal{E}$ are indeed \emph{functions} like the elements of $\mathcal{V}$, the elements of $\mathcal{D}$ are instead finite collections of parameters and hence can be modeled as vectors (unlike the elements of $\mathcal{U}$, which are indeed functions). We now describe our formulation in detail; this formulation is visualized in Figure \ref{fig:ol-pic}.

As mentioned previously, let $\mathcal D$ denote the \emph{design space} of tendon--driven continuum robots (TDCRs). Each element ${\bf d}\in\mathcal D$ encodes a robot design with $N_t$ tendons consisting of several parameters:  
\begin{align}
{\bf d} = \{ \{\boldsymbol\rho_i\,\phi_i,\tau_i\}_{i=1}^{N_t},r, L,E \},
\end{align}
where $\boldsymbol\rho_i$ is once again the cross-sectional offset of each tendon, $\phi_i$ is the tendon path, $\tau_i$ is the tension on each tendon, $L$ is the backbone length, $r$ its radius, and $E$ its Young's modulus. As implied above, we consider tendon tensions as the only actuation input for this work. For each ${\bf d}$ the Cosserat model defines an \emph{equilibrium configuration}
\begin{equation*}\label{eq:main-eq}
\mathcal E({\bf d},s) := \{ (\mathbf r_{\bf d}(s), R_{\bf d}(s) ) \},
\end{equation*}
where $\mathbf r_{\bf d}:[0,L]\to\mathbb R^3$ is the centerline and $R_{\bf d}:[0,L]\to SO(3)$ the frame field-- both for a specific design ${\bf d}$ -- satisfying the rod balance equations with tendon loads~\cite{Antman2005Problems,Rao2021HowToModelTDCR}. This leads naturally to the definition of the infinite-dimensional equilibrium operator $\mathcal{G}$, which is a mapping
\begin{equation}
\mathcal{G} : {\bf d} \longmapsto (\mathbf r(\cdot), R(\cdot)), 
\qquad \mathcal G: \mathcal D \to \mathcal X, \label{eq:equilibrium-operator}
\end{equation}
with $\mathcal X$ the space of equilibrium configurations. Interestingly, the problem also directly admits a simpler alternative definition of the space $\mathcal{E}$ which we label $\mathcal{E}'$. It is possible to simply define the equilibrium configurations purely in terms of the tendon positions so that
\begin{align}\label{eq:alt-eq}
    \mathcal{E}'({\bf d},s) = \{{\bf r}_i({\bf d},s) \}_{i=1}^{N_t}.
\end{align}
Let $\mathcal{X}'$ be the space of all such equilibrium configurations now defined as the set of all $\mathcal{E}'$ for every ${\bf d}$. Then, naturally, we have the alternative definition of the equilibrium operator as $\mathcal{G}': \mathcal{D} \to \mathcal{X}'$. We emphasize that $\mathcal{E}$ is a function that consumes a design ${\bf d}$ and an arclength $s$ and outputs a vector in $\mathbb{R}^3$ and a $3 \times 3$ rotation matrix, while $\mathcal{E}'$ is a function that consumes the same inputs but instead outputs $N_t$ vectors in $\mathbb{R}^3$. From the mathematical perspective, there is a strict one-to-one mapping between $\mathcal{E}$ and $\mathcal{E}'$ for each practical design ${\bf d}$, and consequently $\mathcal{G}$ and $\mathcal{G}'$ are isomorphic. 

We now discuss realizations of these operators through two neural operator architectures: DeepONets and FNOs. It is important to note that these operators could also be learned from actual data rather than Cosserat rod simulations.

\subsection{Why Operator Learning}

The Cosserat rod equations define a nonlinear boundary-value problem in arclength, and their solution induces a nonlinear solution operator. In the classical setting, this operator maps prescribed input functions and boundary conditions to rod configurations and internal force and moment fields. In our setting, we instead parameterize the inputs by the robot design variables together with the distributed loading, and seek the corresponding equilibrium state variables as output functions. This still defines a well-posed operator mapping from a finite-dimensional design/load space into an appropriate function space of rod states. Because repeated evaluation of this operator requires the numerical solution of a nonlinear system, it can become computationally expensive in design, control, and optimization workflows. This motivates the use of operator-learning methods to construct a data-driven surrogate of the solution operator, with the goal of substantially reducing evaluation cost while preserving the underlying input–output structure of the problem.

\subsection{DeepONet Architectures}
\label{sec:don}
\begin{figure}[!htpb]
\centering
\begin{tikzpicture}[>=Latex,thick]

  \node[draw,rounded corners,fill=blue!6,draw=blue!55,minimum width=3.2cm,minimum height=1.0cm] (branch)
    {$\beta_{\theta_1} (\mathbf d)$};
  \node[draw,rounded corners,fill=orange!25,draw=orange,minimum width=3.2cm,minimum height=1.0cm,
        below=2.2cm of branch] (trunk)
    {$T_{\theta_2}(s)$};

  \path let \p1=(branch.south), \p2=(trunk.north) in
    node[circle,draw,fill = blue!20!orange!, minimum size=9mm] (fuse) at ($(\p1)!.5!(\p2)$) {$\langle\cdot,\cdot\rangle$};

  \draw[->] (branch.south) -- (fuse.north);
  \draw[->] (trunk.north) -- (fuse.south);

  \node[draw,rounded corners,fill=yellow,minimum width=3.2cm,minimum height=0.9cm,
        right=2.5cm of fuse, yshift=1.2cm] (rout)
    {$\mathbf r_{\mathbf d}(s)\in\mathbb R^3$};
  \node[draw,rounded corners,fill=blue!20,minimum width=3.2cm,minimum height=1.05cm,
        right=2.5cm of fuse, yshift=-1.2cm] (Rout)
    {\shortstack{$R_{\mathbf d}(s)\in SO(3)$ \\ \scriptsize $R^\top R=I,\ \det R=1$}};

  \draw[->] (fuse) -- (rout.west);
  \draw[->] (fuse) -- (Rout.west);

\end{tikzpicture}
\vspace{-0.6em}
\caption{DeepONet: branch $\beta_{\theta_1}(\mathbf d)$ and trunk $T_{\theta_2}(s)$ fuse via an inner product $\inner{.}{.}$ to produce
$\mathcal G_{\mathrm{DON}}(\mathbf d,s,\theta)=\big(\mathbf r_{\mathbf d}(s),R_{\mathbf d}(s)\big)$ from \eqref{eq:don1}. $\mathcal{G}'_{\rm DON}$ instead outputs $N_t$ tendon position vectors.}
\label{fig:don}
\end{figure}
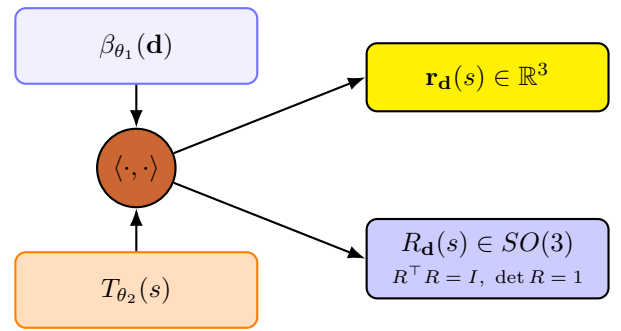
We present two DeepONet architectures, $\mathcal{G}_{\rm DON}$ and $\mathcal{G}'_{\rm DON}$, that learn $\mathcal{G}$ and $\mathcal{G}'$ respectively. These architectures use the same inputs, but differ in their outputs: $\mathcal{G}_{\rm DON}$ outputs in $\mathcal{X}$ for each design ${\bf d}$ (a backbone vector in $\mathbb{R}^3$ and a $3 \times 3$ rotation matrix for each $s$), while $\mathcal{G}'_{\rm DON}$ outputs in $\mathcal{X}'$ ($N_t$ vectors in $\mathbb{R}^3$ for each $s$). $\mathcal{G}_{\rm DON}$ is given explicitly by
\begin{align}\label{eq:don1}
    \mathcal{G}_{\rm DON}({\bf d},s,\theta) =  \inner{\beta_{\theta_1}({\bf d})}{T_{\theta_2}(s)},   
\end{align}
where $\beta_{\theta_1}({\bf d})$ is a (vector-valued) neural network with trainable parameters $\theta_1$ (the ``branch net'') that encodes the design space $\mathcal{D}$; $T_{\theta_2}(s)$ is another (vector-valued) neural network with trainable parameters $\theta_2$ that spans equilibrium configuration space $\mathcal{X}$; $\theta = \{\theta_1, \theta_2\}$; and $\inner{.}{.}$ represents the $\ell_2$ inner product. The branch and trunk are trained jointly to minimize some loss function; see Figure \ref{fig:don} for an illustration. $\mathcal{G}'_{\rm DON}$ has the same high level architecture as well, which we write as
\begin{align}\label{eq:don2}
    \mathcal{G}'_{\rm DON}({\bf d},s,\theta') =  \inner{\beta'_{\theta'_1}({\bf d})}{T'_{\theta'_2}(s)}.
\end{align}
Note the primes over the symbols to distinguish \eqref{eq:don2} from  \eqref{eq:don1}. 

\subsection{FNO Architectures}
\label{sec:fno}
\begin{figure}[!htpb]
\centering
\begin{tikzpicture}[>=Latex,thick,node distance=3.0mm]
  
  \node[draw,rounded corners,fill=blue!6,draw=blue!55,minimum width=7mm,minimum height=6mm,inner sep=1pt]
    (d) {\scriptsize $\mathbf d$};

  \node[draw,rounded corners,fill=gray!15,minimum width=7mm,minimum height=6mm,inner sep=1pt,
        right=of d] (L) {\scriptsize $\mathcal L$};
  \draw[->] (d) -- (L);

  \node[draw,rounded corners,fill=blue!20!orange!,minimum width=7mm,minimum height=6mm,inner sep=1pt,
        right=of L] (A1) {\scriptsize $\mathcal A_1$};
  \node[minimum width=5mm,inner sep=0pt,right=of A1] (dots) {\scriptsize $\cdots$};
  \node[draw,rounded corners,fill=blue!20!orange!,minimum width=7mm,minimum height=6mm,inner sep=1pt,
        right=of dots] (AL) {\scriptsize $\mathcal A_L$};
  \node[draw,rounded corners,fill=orange!25,draw=orange,minimum width=7mm,minimum height=6mm,inner sep=1pt,
        right=of AL] (P) {\scriptsize $\mathcal P$};

  \draw[->] (L) -- (A1);
  \draw[->] (A1) -- (dots);
  \draw[->] (dots) -- (AL);
  \draw[->] (AL) -- (P);

  \node[above=1.2mm of A1] {\scriptsize $\sigma$};
  \node[above=1.2mm of AL] {\scriptsize $\sigma$};

  \node[draw,rounded corners,fill=yellow,minimum width=16mm,minimum height=7mm,inner sep=1pt,
        right=7mm of P,yshift=6.5mm] (rout) {\scriptsize $\mathbf r_{\mathbf d}(s)\in\mathbb R^3$};
  \node[draw,rounded corners,fill=blue!20,minimum width=16mm,minimum height=7mm,inner sep=1pt,
        right=7mm of P,yshift=-6.5mm] (Rout) {\scriptsize $R_{\mathbf d}(s)\in SO(3)$};

  \draw[->] (P.east) -- (rout.west);
  \draw[->] (P.east) -- (Rout.west);

\end{tikzpicture}
\vspace{-0.6em}
\caption{FNO: the input design ${\bf d}$ is lifted and passed through activated Fourier layers \eqref{eq:fno_int} before it is projected using a DNN down to the equilibrium configuration, giving $\mathcal{G}_{\rm FNO}$ in \eqref{eq:fno1}. $\mathcal{G}'_{\rm FNO}$ instead outputs $N_t$ tendon position vectors.}
\label{fig:fno-one-column}
\end{figure}
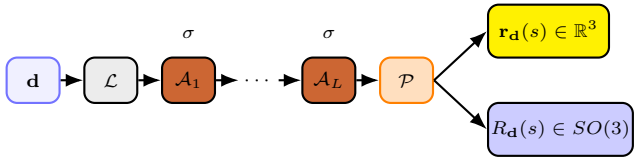
We also developed two FNO architectures, $\mathcal{G}_{\rm FNO}$ and $\mathcal{G}'_{\rm FNO}$, that learn $\mathcal{G}$ and $\mathcal{G}'$ respectively. FNO architectures differ significantly from DeepONets in that they are constructed by compositions of kernel-based integral operators parametrized by the fast Fourier transform (FFT), rather than from ``vanilla'' DNNs. More precisely, the FNO $\mathcal{G}_{\rm FNO}$ can be written as
\begin{align}\label{eq:fno1}
    \mathcal{G}_{\rm FNO}({\bf d}) = \mathcal{P} \circ \mathcal{A}_L \circ \sigma \circ \ldots \circ \sigma \circ \mathcal{A}_1 \circ \mathcal{L} \circ {\bf d},
\end{align}
where $\circ$ indicates composition; $\mathcal{L}$ is a linear layer that takes an $n \times |{\bf d}|$ matrix containing the tensor product of $n$ arclength samples along the backbone with the design vector ${\bf d}$ and outputs an $n \times p$ matrix where $p$ is the ``channel dimension''; $\sigma$ is a nonlinear (elementwise) activation; $\mathcal{P}$ is a standard DNN that projects (undoes $\mathcal{L}$); and $\mathcal{A}_k$ is a ``Fourier layer'', the functional equivalent of a standard DNN's affine transform but with an additional residual connection. To see the structure of $\mathcal{A}$, we first define the argument ${\bf q} = ({\bf d},s)$, then write
\begin{align}\label{eq:fno_int}
    \mathcal{A} ({\bf u}({\bf q})) = \int_{D_{\bf q}} K({\bf q},{\bf q}') {\bf u}({\bf q}') d{\bf q}' + W \underbar{{\bf u}},
\end{align}
where $\underbar{{\bf u}}$ is the vector obtained by sampling the input function to the layer (${\bf u}({\bf q})$) on the tensor-product of the design and arclength grids. The integral in \eqref{eq:fno_int} is handled via the FFT in that entries of the Gramian of $K$ are learned directly in frequency space; we retain the 5 lowest frequency Fourier modes (we experimented with others, but found this performed the best). As in the DeepONet case, we also created a variant $\mathcal{G}'_{\rm FNO}$ to directly output tendon equilibrium positions (i.e., to learn $\mathcal{G}'$); again as in the DeepONet case, $\mathcal{G}_{\rm FNO}$ is trained to output a backbone position and two columns of the rotation matrix for each $s$. 

\subsection{Loss functions}
\label{sec:loss}
We now discuss the loss functions used to train our neural operators. For all our models, we generate the same training dataset of $N$ training pairs $\{ {\bf d}_j, \mathcal{E}'({\bf d}_j) \}_{j=1}^{N}$ (design-- tendon equilibrium position) using a Cosserat rod simulator, where $\mathcal{E}'$ is given by \eqref{eq:alt-eq}. However, since the outputs of both types of models ($\mathcal{G}$ and $\mathcal{G}'$) are different, we use different loss functions for each. Both the DeepONet and FNO variants $\mathcal{G}_{*}$ that approximate $\mathcal{G}$ output a backbone position and rotation matrix, so the loss function is defined as
\begin{align}\label{eq:loss1}
    e({\bf d},s) = \sum\limits_{i=1}^{N_t} \| {\bf r}_i({\bf d},s) - \left( \tilde{\bf r}_{\bf d}(s) + \tilde{R}_{\bf d}(s)\boldsymbol\rho_i \right) \|^2_2,
\end{align}
where $\tilde{\bf r}_{\bf d}(s)$ is the backbone position \emph{estimated} by the neural operator $\mathcal{G}_{*}$ and $\tilde{R}_{\bf d}(s)$ is the estimated rotation matrix. As mentioned previously, we ensure the columns of this $3 \times 3$ rotation matrix are orthonormal by using the Gram-Schmidt process to orthonormalize the two columns output by $\mathcal{G}_{*}$ and to generate a third column as well; we do this during both training and inference. On the other hand, to train neural operators $\mathcal{G}'_{*}$, we instead define the loss
\begin{align}\label{eq:loss2}
    e'({\bf d},s) = \sum\limits_{i=1}^{N_t} \| {\bf r}_i({\bf d},s) - \tilde{{\bf r}}_i({\bf d},s) \|^2_2,
\end{align}
where $\tilde{{\bf r}}_i({\bf d},s)$ are now the tendon equilibrium positons directly predicted by the neural operators $\mathcal{G}'_{*}$. Both $e$ and $e'$ are averaged over all $N$ training designs ${\bf d}$ and evaluated on an $n$ point grid for the variable $s \in [0,L]$ to obtain fully discrete losses.



\subsection{More Architectural Details}
\label{sec:impl-details}
We next describe the practical implementation of the DeepONet and FNO architectures evaluated in this work.

\paragraph{DeepONet architectures} For all our DeepONets, we use the ``stacked'' vanilla-DeepONet architecture (single branch rather than many) with multilayer perceptrons (MLPs) as both the branch $\beta$ and trunk $T$. The branch network has five layers with hidden dimension $64$, using $\tanh$ activations after each layer except the output (which is unactivated, as is customary). The trunk network again has five layers and a hidden dimension of 128, once again with $\tanh$ activations. For the DeepONet $\mathcal{G}'_{\rm DON}$, the output dimension is $12p$ for vanilla DeepONet (corresponding to three spatial coordinates for each of the four tendons), while the output dimension for the DeepONet $\mathcal{G}_{\rm DON}$ is $9p$ (three coordinates for the centerline offset and six values used to construct the rotation matrix via the Gram–Schmidt method). In our experiments, we found that $p=100$ performed the best in both architectures. This results in the vanilla DeepONet $\mathcal{G}'_{\rm DON}$ having 219,904 trainable parameters; $\mathcal{G}_{\rm DON}$ has 205,668 parameters.

\paragraph{FNO architectures} For FNOs, evaluation points are taken along the arclength of the centerline as an equispaced grid on $[0,L]$, concatenated with the model inputs. This two-dimensional input array is lifted to the channel dimension $p=128$ and passed through five Fourier layers, each retaining five modes with ReLU activations in between. A final linear layer maps the output to the target dimension of 12 for $\mathcal{G}'_{\rm FNO}$ (three coordinates for each of four tendons) and 9 for $\mathcal{G}_{\rm FNO}$ (three coordinates for the centerline offset and six values for the rotation matrix via Gram–Schmidt). These architectural choices results in $\mathcal{G}'_{\rm FNO}$ having 168,844 trainable parameters and $\mathcal{G}_{\rm FNO}$ having 168,457 parameters.
\subsection{Training}
\label{sec:training}

\subsubsection{Data generation and preprocessing}
\label{sec:data-gen}
We briefly describe our data generation and preprocessing. Input parameters were uniformly sampled over predefined ranges; these are shown in Table \ref{tab:inputRangeTable}) to obtain $N$ designs. For each design ${\bf d}$, a Runge–Kutta solver was used to solve for $N_t =4$ tendon positions in $\mathbb{R}^3$. Each of the $N$ equilibrium configurations thus consists of 12 values (three spatial coordinates for four tendons) recorded at $n=42$ equispaced points along the robot centerline/backbone. The corresponding arclength values were also stored and used as inputs to the neural operators. 

This dataset was then  partitioned into training and test sets by holding out 20\% of the samples for testing, with the remaining 80\% used for training. For our largest dataset, we generate 100,000 training examples, using $N=80,000$ for training and 20,000 for testing. Models are trained using the full 80,000 examples unless stated otherwise, but the data is put into batches of size 5000 to allow the data to fit into GPU memory.
\begin{table}[h!]
\caption{Ranges of input parameters used to generate training and test data. \label{tab:inputRangeTable}} 
    \centering        
    \begin{tabular}{lcc}
        \toprule
        \textbf{Parameter} & \textbf{Symbol} & \textbf{Range} \\
        \midrule
        Tendon tensions & $\tau_i$    & $[0.0, 5]$ N \\
        Length of robot         & $L$   & $[0.1, 0.35]$ m \\
        Tendon pitches & $\phi_i$   & $[-20, 20]$ rads/m \\
        Tendon offset           & $\rho_i$   & $[0.005, 0.01]$ m \\
        Young's modulus         & $E$   & $[15.5, 45.5] \times 10^9$ Pa \\
        Backbone radius         & $r$   & $[.0005, .0015]$        
    \end{tabular}   
\end{table}
As seen in Table \ref{tab:inputRangeTable} our parameters have several magnitudes of difference between them. To mitigate the risk of a single parameter dominating the others, we regularized our inputs by dividing or multiplying certain parameters to ensure that they were within at most one order of magnitude from the others (e.g. dividing the Young's modulus by $10^9$ to reduce its range to [15.5, 46.5]). This allowed us to keep our values from drowning out our smaller values while keeping the feature space of the parameters linear.

\subsubsection{Training procedure}
\label{sec:train-deets}
Our training procedure uses \eqref{eq:loss1} and \eqref{eq:loss2} sampled at $n=42$ arclength locations and $N$ training examples as our discrete loss. We update all model parameters using the Adam optimizer~\cite{kingma2017adammethodstochasticoptimization}. The learning rate is scheduled using the cyclical cosine annealing scheme~\cite{loshchilov2016sgdr}. Our schedule uses an initial learning rate of $10^{-4}$, a warm up fraction of $0.3$, a peak value of $3 \times 10^{-3}$, and end value of $5 \times 10^{-6}$, with 4 cycles and a $\gamma$ of 0.7. We initialize this schedule assuming we will train for 100,000 epochs. However, because we stop training once the $\ell_2$ training error converges, only the first part of the schedule is typically seen. These convergences occurred after around 20000 epochs for the DeepONets and around 1500 epochs for the FNOs.
\section{Results}
\label{sec:results}

\subsection{Preliminaries}
\label{sec:prelim}
For our results section, we represent the effectiveness of our models by measuring their relative $l_2$ error on generalization, $e_{\ell_2}$, defined by:
\begin{align}\label{eq:l2Error}
     e_{\ell_2} =   \frac{\| y - \hat{y} \|_{2}}{\| y \|_{2}},
\end{align}
where y is the ground truth vector, and $\hat{y}$ is the model's prediction of y. The accuracy of the model can be computed straightforwardly as $(1 - e_{\ell_2}) * 100\%$ to give us a percent accuracy of our model. For each result in this section. We train the models over 3 different seeds and report the average of the relative errors for our results. For the remainder of this section, we will refer to each of our models with the following labels:
\begin{itemize}
    \item $\mathcal{G}_{\rm DON}$ as ``DeepONetPose''; $\mathcal{G}'_{\rm DON}$ as ``DeepONet''.
    \item $\mathcal{G}_{\rm FNO}$ as ``FNOPose''; $\mathcal{G}'_{\rm DON}$ as ``FNO''.
\end{itemize}
The rationale for this naming scheme is that the ``Pose'' variants explicitly output a ``pose'': a backbone and a local orthonormal basis to describe the offset to the tendons. All experiments were run on an Nvidia RTX 4090 GPU with code written using the JAX python library.

\subsection{Generalization Errors}
\label{sec:conv}
\begin{figure}[!htpb]
    \centering
    \includegraphics[width=1.0\linewidth]{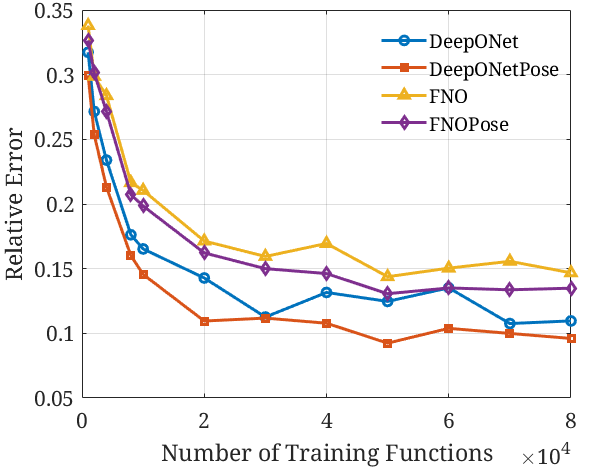}
    \caption{Decay in generalization error as a function of number of training pairs $N$. We increase the number of training designs and the corresponding equilibrium configurations for all four models and compute errors according to \eqref{eq:l2Error}.}
    \label{fig:L2vN}
\end{figure}
First, we explored the generalization error as the number of design-equilibrium pairs $\{{\bf d}_i, \mathcal{E}({\bf d}_i \}_{i=1}^N$ is increased by training each model to (training) convergence. This type of convergence study (error as a function of $N$) is common in the neural operator literature. The idea is that the limit $N \to \infty$ corresponds to recovering the infinite-dimensional operator; in this setting, an absence of error decay would be concerning. The results are shown in Figure \ref{fig:L2vN}.

Figure \ref{fig:L2vN} shows that the accuracy of the model improves rapidly as $N$ increases, but the rate of decrease of the error begins to slow down around $N=20,000$ training pairs with the minimum error achieved for different models at different values of $N$. This is likely due to the fact that the model capacity has been reached; further decrease in error will likely require wider or deeper networks, or perhaps a higher value of $p$ (for both DeepONets and FNOs), or possibly more advanced architectures (such as those based on transformers and attention, for instance). Another trend that can be clearly seen is that DeepONets consistently achieve lower relative errors than FNOs. In addition, the ``Pose'' models (which output backbone positions and rotation matrices) outperform the baseline counterparts. The specific number of training functions required for a problem will in general depend on the ranges of the input design. 

\subsection{Model Parsimony}
\label{sec:dropout}
\begin{figure}[!htpb]
    \centering
    \includegraphics[width=1.0\linewidth]{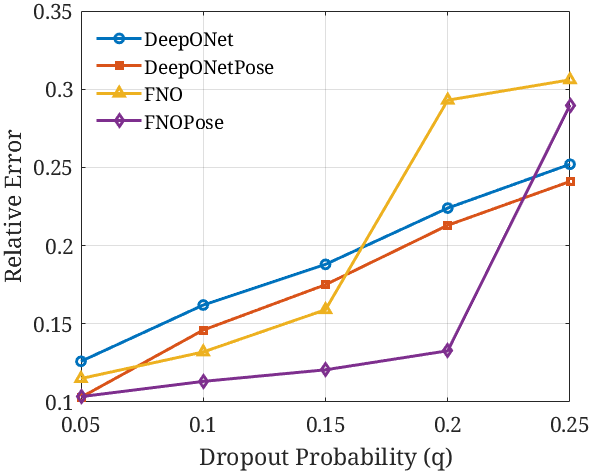}
    \caption{The effect of dropout on generalization errors for $N=80,000$. We see that dropout actually worsens the generalization errors across the board linearly, though it appears to aid the FNOPose model to a certain extent.}
    \label{fig:L2vDrop}
\end{figure}
Neural operators typically have a large number of trainable parameters (as do our models). Naturally, it is reasonable to ask if such high parameter counts are necessary for learning the map from the design space to the equilibrium position space. To explore this question, we added dropout regularization to our training procedure by inserting dropout layers between the layers in our models. Since dropout is the procedure of zeroing out neuron weights randomly,  we use this test as a method of discerning if our models are overparametrized or merely sufficiently parameterized. For the DeepONet, we inserted dropout layers between every layer in the branch and trunk MLPs; for the FNOs, we included dropout between the Fourier layers. We then varied the dropout probability $q$ and trained our models. 

The results are shown in Figure \ref{fig:L2vDrop}, which surprisingly shows that increasing $q$ negatively impacts the relative error (in a linear fashion) even with small probabilities of zeroing out neurons. This overall behavior is indicative that our models appear to be fully utilizing all of their trainable parameters. Interestingly, there is a subtle difference between the DeepONet and FNO models: the DeepONet errors increases linearly with $q$, growing quicker than the FNO errors for small dropouts ($q \leq 0.15$); the latter grow sublinearly. For sufficiently large $q$, however, the FNO errors shoot up to be greater than the DeepONets. The FNO models seem to benefit slightly more from dropout; this is potentially due to dropout helping smooth away Gibbs' oscillations caused by the FFT (though this is merely speculation on our part). We observe that the ``Pose'' models and their counterparts seem to behave similarly, though the error in the FNOPose model does shoot up a slightly later $q$ than its counterpart.

\subsection{Out-of-Distribution Predictions}
\label{sec:ood}
\begin{figure}[!htpb]
    \centering
    \includegraphics[width=1\linewidth]{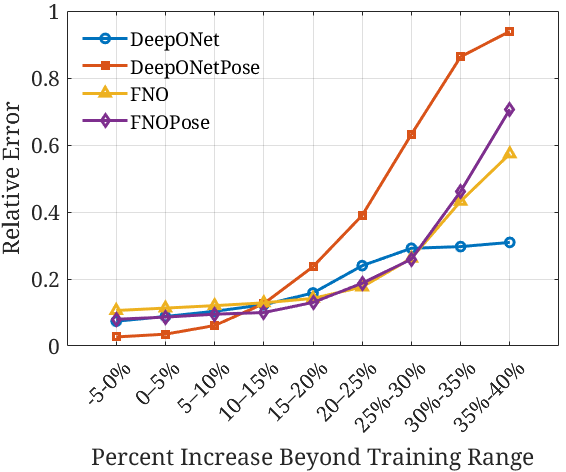}
    \caption{Generalization errors for designs that are outside of the training distribution for models trained on $N = 80,000$ training pairs. The x-axis shows the percentage by which the design parameters are outside the training range.}
    \label{fig:l2OOD}
\end{figure}
We also investigated the ability of our models to extrapolate to out-of-distribution (OOD) inputs. Neural operators are widely regarded as capable of such extrapolation. 

To generate data for this OOD extrapolation test, we chose design parameters of varying ranges beyond our training range. We first created a set of bins, each with an upper and lower percent; specifically, we chose our bins to have a width of 5\% and created four bins ranging the top end from 5\%-20\%. For each of these bins, we generated 1000 sets of random percentages, one random value for each input, within that bin's range. Then, we took the ranges from Table \ref{tab:inputRangeTable}, added the percentage to the right end of that range, and created parameters with those value. We then fed the designs composed of these new OOD inputs through the same Cosserat rod solver to obtain the corresponding equilibrium configurations. Finally, we fed these OOD pairs through each trained model and computed relative errors between the model outputs and the solver outputs. These errors are shown in Figure \ref{fig:l2OOD}.

Figure \ref{fig:l2OOD} shows that our model accuracy falls off once we leave our training range, but with a strange caveat that small deviations actually improve the accuracy over the baseline! For instance, the 0-5\% increase actually \emph{improved} the accuracy of our model, in some cases to 97\% (over the baseline 90\%). A careful inspection of the model parameters reveals the true reason for this improvement: increasing the ranges of our parameters can in some cases make the TDCR backbones stiffer, thereby reducing their range of motion and making the equilibrium configuration easier to predict.

Our first data point for each model shows the in-distribution accuracy of our models (the -5\% to 0\% bin), once again showing that DeepONets are more accurate than FNOs, while the ``Pose'' models outperform their counterparts. The DeepONets errors appear to grow at a greater rate then the FNOs, with the DeepONetPose model's error increasing exponentially until it becomes the least accurate model. We similarly see the FNO and FNOPose model errors increasing exponentially, with the FNOPose overtaking the FNO model at around 20\%, though not as quickly as the DeepONetPose model. Surprisingly, the vanilla-DeepONet begins to show this same behavior, but then flattens out and becomes the most accurate model at high percentages, with very reasonable errors. It seems that the ``Pose'' models (which output a backbone and a rotation matrix) extrapolate worse than those without. This makes some sense as deviations in the columns of the rotation matrices could be amplified through the Gram-Schmidt method, leading to less reliable predictions. In the future, we will explore forming a basis for $SO(3)$ and encoding that into the architectures directly. 

In general, the OOD tests show that the actual physics governing the TDCR can dictate the generalization capabilities of the neural operator surrogates. The OOD performance can likely be improved by either incorporating Cosserat rod physics into the models as a soft loss term or designing neural operators that architecturally respect the Cosserat model. We will explore these avenues in future work.

\subsection{Timings}
\label{sec:inferenceTimings}
\begin{table}[!htpb]
\caption{Model training times as a function of $N$.\label{tab:trainingTimes}} 
    \centering        
    \begin{tabular}{lccc}
        \toprule
        \textbf{Model} & $N=1$ & $N=1000$ & $N=80,000$ \\
        \midrule
        DeepONet            & 149.4 s    & 150.6 s  &   108 min \\
        DeepONetPose        & 144.2 s    & 148.3 s  &   106 min \\
        FNO                 & 13.66 s    & 14.66 s  &   15.5 min \\
        FNOPose             & 14.56 s    & 15.28 s  &   15.8 min \\   
    \end{tabular}   
\end{table}

\begin{table}[!htpb]
\caption{Model inference times.\label{tab:inferenceTimes}} 
    \centering        
    \begin{tabular}{lccc}
        \toprule
        \textbf{Model} & \textbf{1 design} & \textbf{1k designs} & \textbf{80k designs} \\
        \midrule
        DeepONet            & .0019 s    & .0022 s  &   .0878 s  \\
        DeepONetPose        & .0017 s    & .0021 s  &   .0756 s \\
        FNO                 & .0015 s    & .0015 s  &   .1997 s \\
        FNOPose             & .0014 s    & .0016 s  &   .2009 s \\    
    \end{tabular}   
\end{table}
Finally, we present timings for all four models on an RTX 4090 GPU. The training times are shown for $N=1$, $N=1000$, and $N=80,000$ in Table \ref{tab:trainingTimes}. While both DeepONets and FNOs take approximately the same time per training epoch, our DeepONets take O(20k) epochs to converge, while our FNOs require only O(2k) epochs for convergence. DeepONets were more accurate on generalization, but FNOs seem to possess a better accuracy-to-training-time tradeoff. It is certainly worth noting that our DeepONets required almost 2 hours to train on the largest $N$ values while the FNOs only required 15 minutes.

Of course, the modern perspective is that training is preprocessing, and that inference times are more important. In Table \ref{tab:inferenceTimes}, we show some inference times for predicting equilibrium configurations from designs using pretrained models. Here, we see that both the DeepONet and FNO models are capable of rapid inference, with DeepONets able to produce equilibria for $80,000$ designs in less than 1/10 of a second, and FNOs in about 2/10 of a second. Given that all models are able to achieve high accuracy (approximately 90\%) on generalization to never-seen-before designs, these results provide strong evidence that our neural operator models are well-suited to accurate and rapid inference. An alternative perspective is that neural operators can be used as an efficient and accurate forward model for real-time tasks involving TDCRs.

\section{conclusion}
\label{sec:conclusion}
In this work, we developed a mathematical framework for casting the problem of learning maps from TDCR designs to their equilibrium positions as an operator learning problem. We then developed four design-agnostic models based on neural operators that can generalize to new designs at runtime. We demonstrated the ability of these models to accurately and rapidly predict robot kinematics on new designs never seen before.

In this work, we demonstrated results for 4-tendon TDCRs. That said, our models are truly design-agnostic; they accept tendon paths, tensions, and offsets from the backbone. Adding more tendons will only require passing additional parameters (the number of tendons) and adding the extra paths, tensions, and offsets into the vector of parameters that determines a design. Further, in this work we designed surrogates based on DeepONets and FNOs. However, our mathematical framework is general and will lend itself well to other neural operators.

This work takes significant steps toward a learned model that is accurate for the spectrum of designs rather than specific to one, enabling faster optimization, control, and planning for these robots.

\bibliographystyle{siam}
\bibliography{refs,adk_tendon_robots}
\end{document}